\documentclass[conference]{IEEEtran}
\IEEEoverridecommandlockouts
\usepackage{cite}
\usepackage{amsmath,amssymb,amsfonts}
\usepackage{algorithmic}
\usepackage{graphicx}
\usepackage{textcomp}
\usepackage{multicol}
\usepackage{xcolor}
\def\BibTeX{{\rm B\kern-.05em{\sc i\kern-.025em b}\kern-.08em
    T\kern-.1667em\lower.7ex\hbox{E}\kern-.125emX}}
\begin{document}

\title{Light-Weighted CNN for Text Classification\\
{\footnotesize }
\thanks{}
}

\author{\IEEEauthorblockN{Ritu Yadav}
\IEEEauthorblockA{\textit{Supervisor: Dr. -Ing. Syed Saqib Bukhari}\\
\textit{Project Report: WS 2018/2019} \\
\textit{Department of Computer Science,}\\
\textit{Technical University of Kaiserslautern}\\
\textit{German Research Center for Artificial Intelligence(DFKI),}\\
Kaiserslautern, Germany \\
er.ritu92@gmail.com, saqib.bukhari@dfki.de}
}

\maketitle

\begin{abstract}
manage the documents. For management, documents are categorized into a specific category, and to do these, most of the organizations use manual labor. In today's automation era, manual efforts on such a task are not justified, and to avoid this, we have so many software out there in the market. However, efficiency and minimal resource consumption is the focal point which is also creating a competition. The categorization of such documents into specified classes by machine provides excellent help. One of categorization technique is text classification using a Convolutional neural network(TextCNN). 

TextCNN uses multiple sizes of filters, as in the case of the inception layer introduced in Googlenet. The network provides good accuracy but causes high memory consumption due to a large number of trainable parameters. As a solution to this problem, we introduced a whole new architecture based on separable convolution. The idea of separable convolution already exists in the field of image classification but not yet introduces to text classification tasks. With the help of this architecture, we can achieve a drastic reduction in trainable parameters.
\\
\end{abstract}


\section{Introduction}
Organizations are reducing paper usage and going online these days, but still, we have few types of organizations such as insurance companies whose large base is operating on physical documents. Physical documents can be handled either manually or using technology. To avoid manual efforts, technical solutions are in place. 
Various machine learning and deep learning algorithms such as support vector machine(SVM), convolutional neural network(CNN), recurrent neural network(RNN), long short term memory(LSTM) is used for classification purposes. To do that, the first documents are converted into digital format by scanning. These scanned documents are images that can be classified further using either image classification method or text classification methods. 

In the text classification domain, Convolutional Neural Networks (CNNs) have recently achieved remarkably strong performance on the practically important task of sentence classification (Kim, 2014; Kalchbrenner et al., 2014; Johnson and Zhang, 2014). However, these models require practitioners to specify an exact model architecture and set accompanying hyperparameters, including the filter region size, regularization parameters, and others. For the scope of the project, we are considering the Text classification convolution neural network(TextCNN) as our base architecture. In TextCNN “Fig. 1”, the first layers embed words into low-dimensional vectors using wordtoVec embedding method. The next layer performs convolutions over the embedded word vectors using filters. Applied convolutions are not homogeneous but use multiple size filters and combine their results after pooling. Max pooling is done on each feature map, where the maximum value from each feature map is forwarded to the next layer forming a long feature vector.  The pooling layer is followed by a fully connected layer. Softmax is used for probabilistic distribution. Drop out mechanism is added for regularization. Also, softmax-cross entropy for loss and RELU activation function is used for adding non-linearity to the network.

CNN text classification architecture[3] gives excellent results with upcoming optimized technology where the high efficiency with fewer resources is in demand. We need to optimize TextCNN as well. The cause of high memory consumption in TextCNN has high number of trainable parameters. The goal is to find a solution by which we can reduce trainable parameters hence memory consumption. We were looking for a new architecture that used some new concept already existing in the image classification world, that can be used for text classification and cause lesser memory consumption. Also, TextCNN does not follow deep neural networks, but an example of a wide network. Hence we cannot play with number of layers.

In the process of finding optimized architecture for document classification, we tried many architectures, and two of them qualify as good solutions to the problem. In the first solution, we tried to alter and optimizes filter sizes. The second solution is a new architecture based on multiple deep learning ideas from recent research. Both solutions are presented in the following sections.

\begin{figure*}[htbp]
\centerline{\includegraphics[width=18cm]{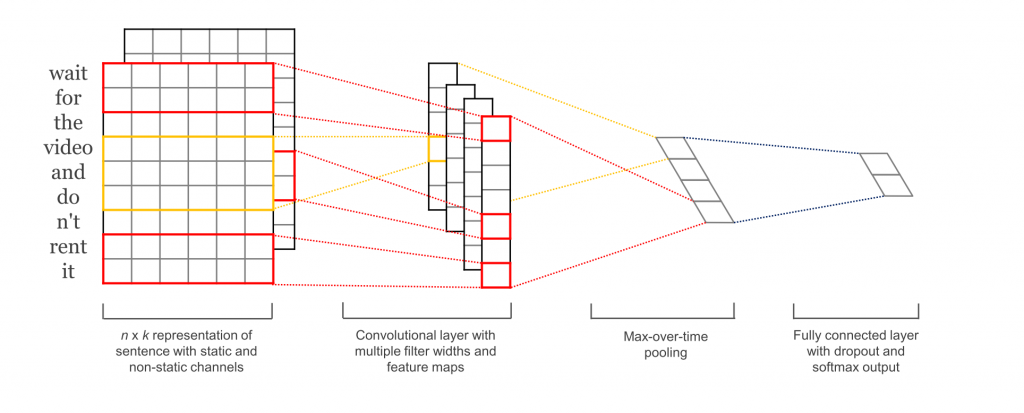}}
\caption{Illustration of a Convolutional Neural Network (CNN) architecture from [8]. n*k representation of a sentence, Convolution layer with multiple filters and feature maps, Max-pooling and Fully connected output layer with dropout and softmax are shown in detail.}
\label{fig}
\end{figure*}

\section{Proposed Solution so far}
\subsection{Optimized Text CNN}
One of the significant reasons behind the large number of trainable parameters in TextCNN is the inappropriate filter size. In TextCNN there are three different size filters 3,4 and 5. Each different size filter is applied 128 times. 
At first, we experimented on the existing TextCNN network. We tried and tested multiple changes to optimize the existing network. The following are the changes established changes that helped in optimization.
First, we reduced the number of filters of each size from 128 to 120. It is tested and verified that it causes no change in accuracy and loss but helps in reducing the number of parameters in an optimized architecture.
Second, we experimented with filter size. We changed the filter size from 3 to filter size 2. Which again helped in reducing the number of trainable parameters. Then all filters of size four were replaced by filters of size 3. The idea behind this filter size change was that the 3x3 size filter is proved efficient in so many neural networks. We utilized the same. There is no change in size for the filters of size 2 and 5. 

To avoid over-fitting, We also added the L2 regularization method. In neural networks, the learning rate of the gradient descent strongly affects performance. This prevents reliable out-of-the-box training of a model on a new problem. Instead of keeping the learning rate the same throughout the training process, we tried learning rate decay for stable and better accuracy. With these changes in architecture, we were able to reduce the number of trainable parameters along with increment in the network's accuracy. 

Further to the above optimization steps in the network, we replaced each filter of size five by two consecutive filters of size 3. The idea behind this change is that two consecutive filters cover the same receptive field[Inception v3 paper] as one filter of size 5. The advantage of using 2 consecutive 3 size filter is that one such filter adds 9 parameters to the network instead of 25 when size 5 filter is used. Also, instead of using 128 filters of each size, 120 filters are used. When all 120 filters are replaced with optimized size, it causes a reduction of a large number of trainable parameters. 
With the above changes, we noticed no further improvements inaccuracy or loss as compared to our base architecture TextCNN.

\subsection{Lightweight CNN}
Lightweight CNN is an architecture proposed to solve the high memory consumption issues. The architecture of the proposed network is visible in the computational graph in “Fig. 2”. The proposed network uses a combination of separable convolution, dilated convolution, and batch normalization in order to reduce memory consumption. All three are explained in detail in the following subsections.

\subsubsection{Separable Convolution}
Separable convolution is the compression technique famous in 3D vision tasks[4] to reduce memory consumption and computation cost. Separable convolution is implemented successfully in “Xception”[1] and “Mobilenet” neural networks.

A depthwise separable convolution “Fig. 3”, commonly called “separable convolution” in deep learning frameworks such as TensorFlow and Keras, consists in a depthwise convolution, i.e., a spatial convolution performed independently over each channel of an input, followed by a pointwise convolution, i.e., a 1x1 convolution, projecting the channels output by the depthwise convolution onto a new channel space. When used in image community, depthwise convolutions as usually implemented (e.g., in TensorFlow) perform channel-wise spatial convolution. Normally the number of filters in convolution is equal to the number of channels in the input image. 

\begin{figure*}[htbp]
\centerline{\includegraphics[width=12cm]{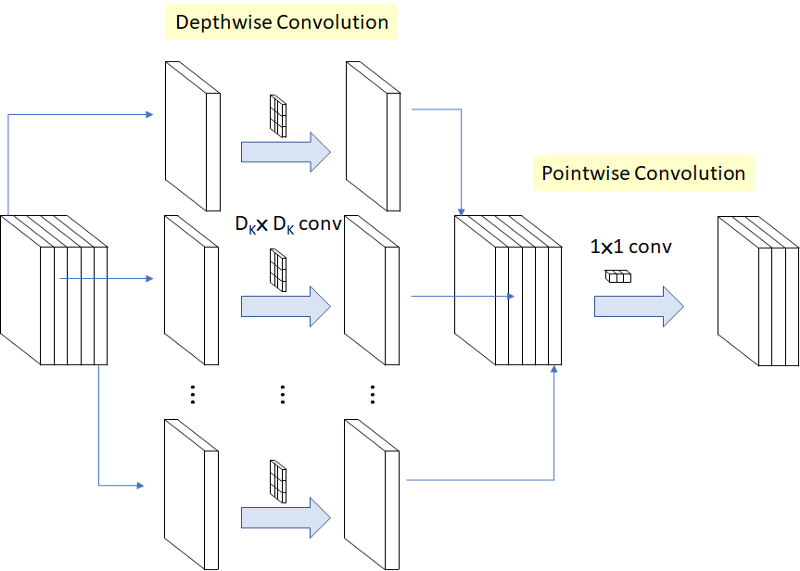}}
\caption{Separable Convolution}
\label{fig}
\end{figure*}
In-text classification, we use embedding, which is 1D So the number of channels is one. This means the entire concept of depthwise convolution cannot be used. Now the idea is to take the benefit of depth convolution and merge this with the modified CNN architecture suggested in the previous section. So we utilize the optimized number and filter size (2,3,5) as proposed in architecture under "Optimized Text CNN: Architecture 1". Also, instead of applying 120 filters, each of size 2,3 and 5(3,3), one filter of each size is applied.

Convolution is then followed by pointwise convolution where 120, 1x1 convolution filters are applied on each individual outputs from the previous layer. 

1x1 convolution is used for two purposes. One is to increase the depth of the network without a potential increase in the number of parameters. As we know, "deep" networks are better than "wide" networks, 1x1 convolution helped in achieving the benefit. The second 1x1 convolution introduces non-linearity to the network when followed immediately by RELU activation. Ultimately, in order to increase the number of non-linear layers without significantly increasing the number of parameters and computations, we can apply a 1x1 kernel and add an activation layer after it. This helps in adding a layer of depth to the network.

\begin{figure*}[htbp]
\centerline{\includegraphics[width=12cm]{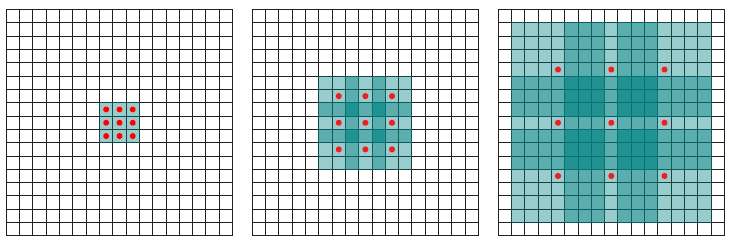}}
\caption{Systematic dilation supports exponential expansion of the receptive field without loss of resolution or coverage. (a) Filter in left figure(F1) is produced from F 0 by a 1-dilated convolution; each element in F1 has a receptive field of 3x3. (b) Filter in middle figure(F2) is produced from F1 by a 2-dilated convolution; each element in F2 has a receptive field of 7x7. (c) Filter in right figure(F3) is produced from F2 by a 4-dilated convolution; each element in F3 has a receptive field of 15x15. The number of parameters associated with each layer is identical. The receptive field grows exponentially while the number of parameters grows linearly.}
\label{fig}
\end{figure*}

\subsubsection{Dilated Convolution}

Dilated convolutions[5], also known as atrous convolutions “Fig. 3” introduce another parameter to convolutional layers called the dilation rate. Dilation rate defines the spacing between the values in a kernel. A 3x3 kernel with a dilation rate of 2 will have the same field of view as a 5x5 kernel, hence introduce 2 extra spacing in the kernel. It can be visualized as taking a 5x5 kernel and deleting every second column and row.  while only using 9 parameters.

Our actual motive behind using dilated convolution is reducing the number of parameters. As we already know that a 3x3 dilated kernel cover receptive field the same as 5x5 kernel. We can see here 3x3 dilated kernel cause 9 parameters, but 5x5 kernel cause 25 parameters to the network. This delivers a wider field of view at the same computational cost. Dilated convolutions are particularly popular in the field of real-time segmentation. It is used when we need a wide field of view but cannot afford the parameter increase due to multiple or larger convolution filters.

In the proposed architecture, we added 3x3 dilated kernel instead of 5x5 kernel. Also, to increase the depth of the network, after the dilated convolutional filter, a 3x3 separable convolutional filter is added. By doing so, the network is benefited with an increase in the depth, more receptive field coverage, and a decrease in the number of parameters.

\subsubsection{Batch Normalization}
Two batch normalization layers were added to the architecture. One is before the first convolutional layer, and another one is before RELU. Bach normalization layers help in accelerating the training processes by reducing the covariate shift. Also, it works as a regularizer.

\subsubsection{Other Specifications}
Instead of RELU, leaky RELU is added to the architecture. Leaky RELU adds non-linearity to the network and eliminate the problem of "dying relu". We are using alpha as 0.1.

\begin{figure*}[htbp]
\centerline{\includegraphics[width=12cm,height=11cm]{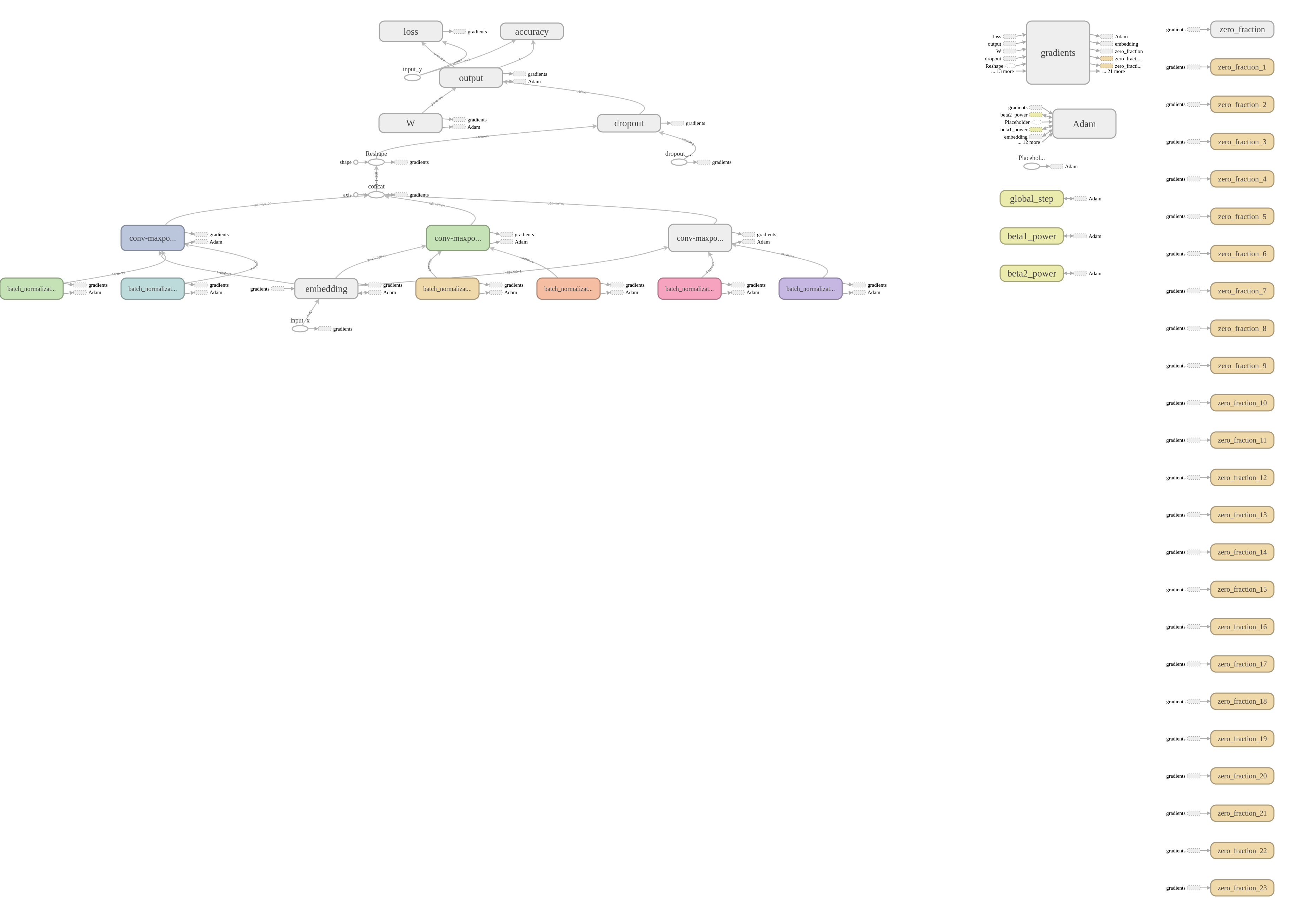}}
\caption{Lightweight text classification CNN Computational graph.}
\label{fig}
\end{figure*}

\begin{figure*}[htbp]
\centerline{\includegraphics[width=12cm,height=12cm]{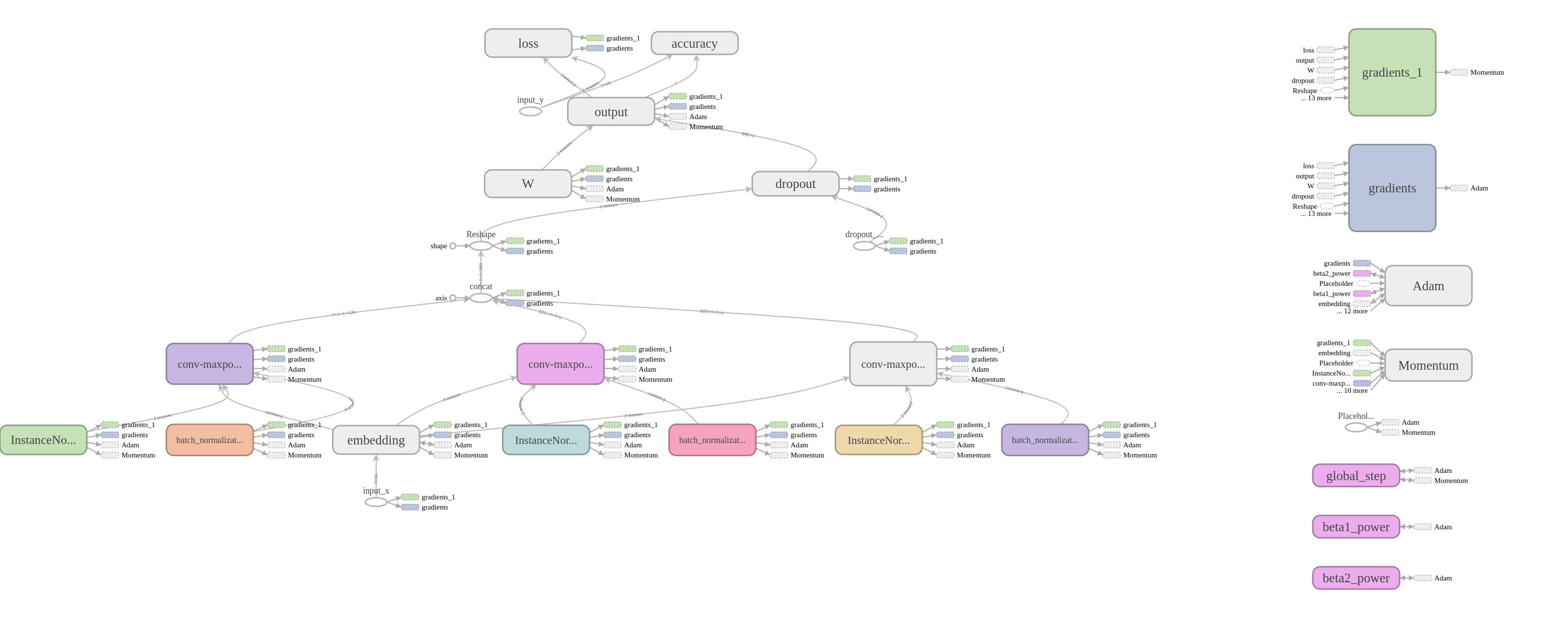}}
\caption{Lightweight text classification CNN with SWAT Computational graph.}
\label{fig}
\end{figure*}
\subsection{DUAL optimizer}
Despite superior training outcomes, adaptive optimization methods such as Adam, Adagrad, or RMSprop have been found to generalize poorly compared to Stochastic gradient descent (SGD). These methods tend to perform well in the initial
portion of training but are outperformed by SGD at later stages of training. So we used a hybrid simple strategy SWATS, which Switches from Adam to SGD when a triggering condition is satisfied.
In this architecture, we tried to utilize the efficiency of SWATS. Two optimizers used were "Adam" and "SGD with momentum". Adam optimizer helps the network to learn fast; hence after a few epochs, we need to slow down the learning process to get stabilized performance. Also, SGD with momentum performs better in a later stage of training, so we tried to get the advantage of the 2 optimizers. We trained the network for few epochs on Adam optimizer and then switched to SGD with momentum for later epochs. The performance of this network is present in the result and evaluation section of the paper.

\section{Dataset}
Datasets used in any study are of utmost importance as they give the intuition of real-life problems. For any study, it is necessary to have a representative dataset because it influences the accuracy of the study hence tells us how well our proposed solution is fit for the problem. 

As we are looking for the solution in document text classification, it is necessary to use a dataset which is well suited for the study and covers a different type of documents encountered for various operations. One such dataset is RVL-CDIP (Ryerson Vision Lab Complex Document Information Processing). RVL-CDIP dataset is publicly available. It consists of 400,000 gray-scale images in 16 classes, with 25,000 images per class. We used two sub-datasets of RVL-CDIP for the study purposes, which are explained in the following sub-sections.

\subsubsection{Tobacco-3482}
The tobacco-3482 dataset consists of images related to tobacco from the media. It has 10 classes and 3482 gray-scale document images combined. The dataset is also publicly available. Classes of Tobacco-3482 are not completely independent of each other. Dataset is complex in nature and consist of overlapping classes. This dataset is a good example of a real-life problem, but at the same time, we are focused on solving the problem with a complex as well as simple classification point of view. Based on the dataset complexity, we will be able to see the difference inaccuracy in the result section.

\subsubsection{Tobacco small-3482}
To solve the inter-dependency of the Tobacco-3482 class documents, we picked up 3 most independent classes from Tobacco-3482 and tested the accuracy of the proposed network. This dataset is not publicly available. We used the confusion matrix “Fig. 4” to get the most independent classes for our study.

\begin{figure}[htbp]
\centerline{\includegraphics[width=75mm]{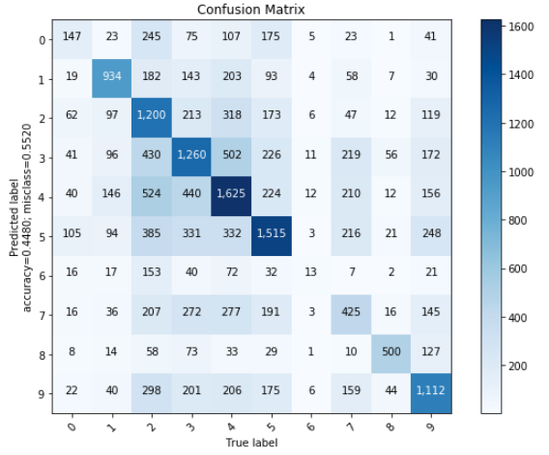}}
\caption{Tobacco-3482 Confusion matrix}
\label{fig}
\end{figure}

\section{Experimental Setup}
In this section we discuss the implementation of the architecture discussed earlier in paper. Implementation is done in Tensorflow.

\subsection{Data Preparation}\label{AA}
Preparation of dataset for any experiment is time taking and crucial task. The Tobacco-3482 dataset contains images with ".tif" extension, and labels are associated as sub-string in the name. For text classification, we need to extract text from these images, which is achieved with the help of tesseract. Tesseract is an optical character recognition(OCR) software that recognizes text characters from images. It takes input as an image file and provides output as a .text file. OCR is a machine-learning technique used to transform images that contain text (e.g., a scan of a document) into actual text content.

Once the text is extracted from the images, our next step is to prepare it for training, validation, and testing data. For that, split the text data into a 9:1 ratio for training and testing data. This training data is then divided into 8:2 ratio for training and validation data. Next, the data is processed by removing the punctuations, line spaces, and others.

\subsection{Implementation and Training }\label{AA}
The first layer defined is the embedding layer, which maps vocabulary word indices into low-dimensional vector representations. It uses a lookup table that we learn from the data, or we can use pre-trained representations here. The next layer performs convolutions over the embedded word vectors using multiple filter sizes. From this point onwards, the following three architectures are different.

In the first architecture, we implemented the base Text CNN with 200 embedding dimensions, 128 number of filters each for size 3,4, and 5. The network is trained on 20 epochs in total.

In the second architecture, we implemented the Lightweight CNN architecture mentioned in the earlier section. First, the output from the embedding layer is normalized using the batch normalization layer. Depthwise convolution is applied to this normalized output. To perform this, 1 filter of each size 2,3,5(3,3) is applied. On top of this Pointwise convolution is applied, which is done by applying 128, 1x1 convolutional filters. Batch normalization is applied to normalize the convolution layer output, which also helped in accelerating the training process.

\begin{figure*}[htbp]
\centerline{\includegraphics[width=12cm,height=8cm]{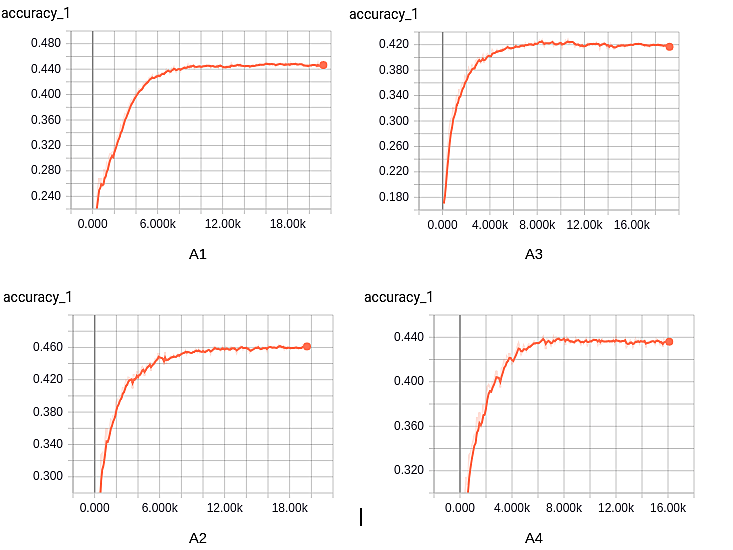}}
\caption{Validation accuracy graphs on Tobacco-3482. A1: accuracy graph of base Text CNN, A2: accuracy graph of Optimized Tex CNN, A3: accuracy graph of Lightweight Text CNN, A4: accuracy graph of Lightweight Text CNN with Dual optimizer. }
\label{fig}
\end{figure*}

\begin{figure*}[htbp]
\centerline{\includegraphics[width=12cm,height=8cm]{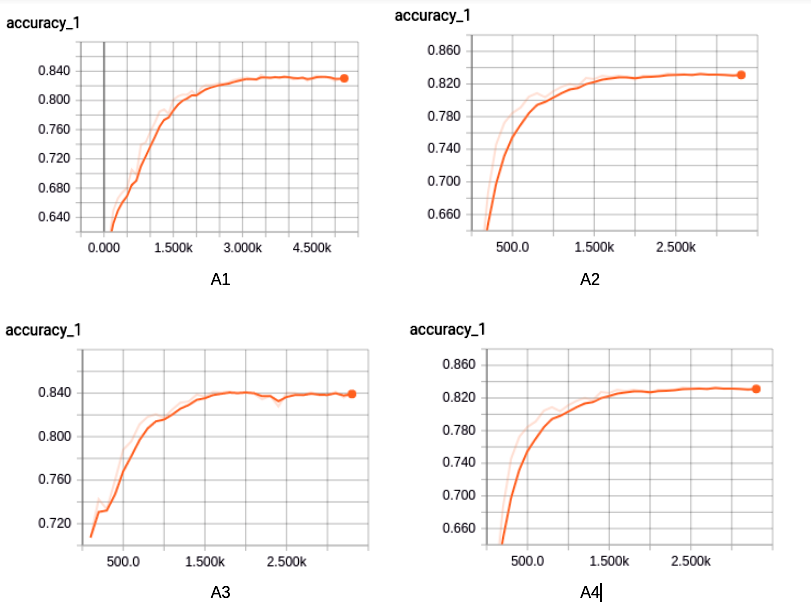}}
\caption{Validation accuracy graphs on Tobacco small-3482. A1: accuracy graph of base Text CNN, A2: accuracy graph of Optimized Tex CNN, A3: accuracy graph of Lightweight Text CNN, A4: accuracy graph of Lightweight Text CNN with Dual optimizer. }
\label{fig}
\end{figure*}

After the above-mentioned operations in different architecture, these networks are followed with a RELU or leaky RELU to add non-linearity, the max pool is applied to extract the most important feature from each output. Results are then concatenated, and the fully connected layer is added. Softmax is applied for probabilistic distribution hence classification. Dropout is used for the network's regularization. During the learning phase of the network, our goal is to reach minimum cost or global minima. In the learning process, the network faces problems like overshoot or plateau. To avoid that, it is a good idea to reduce the learning rate as the network keeps on learning. We used this idea in our proposed architecture. With the training epochs, we reduce the learning rate with the help of a decaying coefficient.

In the third architecture, we tried to get better results by modified the optimizers. Above the second architecture, instead of applying one optimizer, we used 2 different optimizers. The first optimizer used was Adam optimizer, which trains the network very fast. After a few epochs, the optimizer is switched to SGD with momentum. A most important part of this dual optimizer method is to decide the correct optimizer switching condition. We analyzed the accuracy graph, tried multiple attempts to find the right switching condition. This optimizer switching time depends on our model, dataset, and many more factor.

Hyperparameters such as alpha in leaky RELU, optimizer switch steps, learning rate decay coefficients, batch size, l2 regularization value, etc. are tuned by experimenting all networks with different values.
On the Tobacco-3482 dataset, we trained the network for a total of 20 epochs; however, on Tobacco small-3482 dataset training was done for 5 epochs. The idea behind this was that on a small dataset, more training could cause overfitting problems.

\section{Results and Evaluation}
All four networks are compared on the basis of the number of trainable parameters, accuracy, obtained loss, and training time. 
Data for comparison is recorded by the following procedure. In the initial stage of training, the number of trainable parameters is recorded. Evaluation is done after every 100 steps of training. At the end of the training, the final evaluation is done. This is when validation accuracy(in percentage) and loss is recorded. Training time is obtained by recording the start of training and training completion time.

Later the significance of results is checked by using t-testing method with the significance level as 0.05 to reject the null hypothesis.

By evaluating the results, we found that our proposed architecture is compressed as compared to the base TextCNN. With the new model, we are able to get accuracy close to the base Architecture with significantly less loss. With the use of batch normalization, we are able to accelerate the training process, which helped in getting the maximum accuracy in almost half of the training time of base TextCNNN.
Optimizer switching technique,SWAT lead to reduce this training time to one fourth. Dual optimizer use gave more stable results and also improved accuracy to bring it close to that of base TextCNN.
There are so many deep learning concepts that can be applied to new application areas. We used here separable convolution and dilated convolution, which was never experimented in-text classifications before. But when they are explored along which dozens of other methods, we found that not only can they be applied in text classification, but also they gave more optimized results and helped in compressing the model.

\begin{table*}[htbp]
\caption{Comparison of number of trainable parameters,accuracy(in percentage), loss, Memory consumption(Mbps) and training time achieved by different methods on Tobacco-3482 dataset.}
\begin{center}
\begin{tabular}{|c|c|c|c|c|}
\hline
\textbf{}&\multicolumn{4}{|c|}{\textbf{Tobacco-3482}} \\
\cline{2-5} 
\textbf{Method Used} & \textbf{\textit{Trainable Parameters}} & 
\textbf{\textit{Accuracy}} & \textbf{\textit{Loss}}&
\textbf{\textit{Training time}} \\
\hline
Base Text CNN & 16801034 &44.7 & 3.21 & $\sim$2 hour\\
\hline
Optimized Text CNN & 16733570 & 46 & 3.11 & $\sim$2 hour \\
\hline
Lightweight Text CNN & 16496856 & 42 & 1.79 & $\sim$1 hour\\
\hline
Lightweight TextCNN with Dual Optimizer & 16496856 & 43.5 & 1.90 & $\sim$26 min\\
\hline

\end{tabular}
\label{tab1}
\end{center}
\end{table*}

\begin{table*}[htbp]
\caption{Comparison of number of trainable parameters,accuracy(in percentage), loss, Memory consumption(Mbps) and training time achieved by different methods on Tobacco  small-3482 dataset.}
\begin{center}
\begin{tabular}{|c|c|c|c|c|}
\hline
\textbf{}&\multicolumn{4}{|c|}{\textbf{Tobacco small-3482}} \\
\cline{2-5} 
\textbf{Method Used} & \textbf{\textit{Trainable Parameters}} &

\textbf{\textit{Accuracy}} & \textbf{\textit{Loss}}&
\textbf{\textit{Training time}} \\
\hline
Base Text CNN & 6336739 & 83.1 & 0.85 & $\sim$9 min \\
\hline
Optimized Text CNN & 6295043 & 84& 0.60 & $\sim$9 min  \\
\hline
Lightweight Text CNN & 6057603 & 82.5 & 0.65 & $\sim$5 min\\
\hline
Lightweight TextCNN with Dual Optimizer & 6057603 & 83 & 0.54 & $\sim$2 min \\
\hline

\end{tabular}
\label{tab1}
\end{center}
\end{table*}
\section{Conclusion}
In this paper we explained the base Text CNN method and the problem associated with it. We experimented multiple optimization method on existing network to reduce down overall memory consumption. AS a result of our attempts we proposed an optimization solution, which is well explained in the paper with all the results. This solution not only helped in reducing the memory consumption but also increased the model accuracy. 

Further to this We tried multiple new deep learning concepts to built a new architecture for the text classification which can help in reducing the memory consumption. From our experiments we proposed a solution which reduced down the number of trainable parameters by approximately 300000. Along with the memory consumption reduction we achieved other benefits as well. First, loss is reduced by significant amount. Second, training time is by one fourth times.

  \nocite{*}
  \bibliographystyle{plain}
  \bibliography{Refer}
\end{document}